\documentclass[12pt]{article}
\usepackage{amsmath}
\usepackage{amsfonts}
\usepackage{hyperref}
\usepackage{times}
\usepackage{graphicx}
\usepackage{color}
\usepackage{float}
\usepackage{multirow}
\usepackage[authoryear]{natbib}
\usepackage{rotating}
\usepackage{bbm}
\usepackage{latexsym}

\newcommand{\xx}{{\bf x}}
\newcommand{\zz}{{\bf z}}
\newcommand{\yy}{{\bf y}}

\newcommand{\WW}{{\bf W}}
\newcommand{\VV}{{\bf V}}

\newcommand{\bb}{{\bf b}}

\newcommand{\captionfonts}{\normalsize}

\makeatletter  
\long\def\@makecaption#1#2{%
  \vskip\abovecaptionskip
  \sbox\@tempboxa{{\captionfonts #1: #2}}%
  \ifdim \wd\@tempboxa >\hsize
    {\captionfonts #1: #2\par}
  \else
    \hbox to\hsize{\hfil\box\@tempboxa\hfil}%
  \fi
  \vskip\belowcaptionskip}
\makeatother   

\begin{document}
\hspace{13.9cm}1

\ \vspace{20mm}\\

\begin{center}
{\LARGE Correlational Neural Networks\footnote{Work done while the first author was at IIT Madras and IBM Research India.}}
\end{center}
\ \\
{\bf {\large Sarath Chandar}$^{\displaystyle 1}$,  {\large Mitesh M Khapra}$^{\displaystyle 2}$,  {\large Hugo Larochelle}$^{\displaystyle 3}$,\\  {\large Balaraman Ravindran}$^{\displaystyle 4}$   }\\
{$^{\displaystyle 1}$University of Montreal. \texttt{apsarathchandar@gmail.com}}\\
{$^{\displaystyle 2}$IBM Research India. \texttt{mikhapra@in.ibm.com}}\\
{$^{\displaystyle 3}$University of Sherbrooke. \texttt{hugo.larochelle@usherbrooke.ca}}\\
{$^{\displaystyle 4}$Indian Institute of Technology Madras. \texttt{ravi@cse.iitm.ac.in}}\\

%


\thispagestyle{empty}
\markboth{}{NC instructions}
\ \vspace{-0mm}\\
%
\begin{center} {\bf Abstract} \end{center}

Common Representation Learning (CRL), wherein different descriptions (or views) of the data are embedded in a common subspace, is receiving a lot of attention recently. Two popular paradigms here are Canonical Correlation Analysis (CCA) based approaches and Autoencoder (AE) based approaches. CCA based approaches learn a joint representation by maximizing correlation of the views when projected to the common subspace. AE based methods learn a common representation by minimizing the error of reconstructing the two views. Each of these approaches has its own advantages and disadvantages. For example, while CCA based approaches outperform AE based approaches for the task of transfer learning, they are not as scalable as the latter. In this work we propose an AE based approach called Correlational Neural Network (CorrNet), that explicitly maximizes correlation among the views when projected to the common subspace. Through a series of experiments, we demonstrate that the proposed CorrNet is better than the above mentioned approaches with respect to its ability to learn correlated common representations. Further, we employ CorrNet for several cross language tasks and show that the representations learned using CorrNet perform better than the ones learned using other state of the art approaches.

\section{Introduction}

In several real world applications, the data contains more than one view. For example, a movie clip has three views (of different modalities) : audio, video and text/subtitles. However, all the views may not always be available. For example, for many movie clips, audio and video may be available but subtitles may not be available. Recently there has been a lot of interest in learning a common representation for multiple views of the data \citep{ngiam11, KlementievA2012, aps2, aps, dcca, HermannK2014, wang} which can be useful in several downstream applications when some of the views are missing. We consider four applications to motivate the importance of learning common representations: (i) reconstruction of a missing view, (ii) transfer learning, (iii) matching corresponding items across views, and (iv) improving single view performance by using data from other views. 

In the first application, the learned common representations can be used to train a model to reconstruct all the views of the data (akin to autoencoders reconstructing the input view from a hidden representation). Such a model would allow us to reconstruct the subtitles even when only audio/video is available. Now, as an example of transfer learning, consider the case where a profanity detector trained on movie subtitles needs to detect profanities in a movie clip for which only video is available. If a common representation is available for the different views, then such detectors/classifiers can be trained by computing this common representation from the relevant view (subtitles, in the above example). At test time, a common representation can again be computed from the available view (video, in this case) and this representation can be fed to the trained model for prediction. Third, consider the case where items from one view (say, names written using the script of one language) need to be matched to their corresponding items from another view (names written using the script of another language). One way of doing this is to project items from the two views to a common subspace such that the common representations of corresponding items from the two views are correlated. We can then match items across views based on the correlation between their projections. Finally, consider the case where we are interested in learning word representations for a language. If we have access to translations of these words in another language then these translations can provide some context for disambiguation which can lead to learning better word representations. In other words, jointly learning representations for a word in language $L_1$ and its translation in language $L_2$ can lead to better word representations in $L_1$ (see section \ref{sec:bigram}). 

Having motivated the importance of Common Representation Learning (CRL), we now formally define this task. Consider some data ${\cal Z} = \{\zz_i\}_{i=1}^N$ which has two views: $X$ and $Y$. Each data point $\zz_i$ can be represented as a concatenation of these two views : $\zz_i = (\xx_i,\yy_i)$, where $\xx_i$ $\in$ $\mathbb{R}^{d_1}$ and $\yy_i$ $\in$ $\mathbb{R}^{d_2}$. In this work, we are interested in learning two functions, $h_X$ and $h_Y$, such that $h_X(\xx_i) \in \mathbb{R}^k$ and $h_Y(\yy_i) \in \mathbb{R}^k$ are projections of $\xx_i$ and $\yy_i$ respectively in a common subspace ($\mathbb{R}^k$) such that for a given pair $\xx_i$, $\yy_i$ :
\begin{enumerate}
\item $h_X(\xx_i)$ and $h_Y(\yy_i)$ should be highly correlated. 
\item It should be possible to reconstruct $\yy_i$ from $\xx_i$ (through $h_X(\xx_i)$) and vice versa.
\end{enumerate}

Canonical Correlation Analysis (CCA) \citep{cca} is a commonly used tool for learning such common representations for two-view data \citep{Udupa:10,cca_app}. By definition, CCA aims to produce correlated common representations but, it suffers from some drawbacks. 
First, it is not easily scalable to very large datasets.
Of course, there are some approaches which try to make CCA scalable (for example, \citep{fcca}), but such scalability comes at the cost of performance. 
Further, since CCA does not explicitly focus on reconstruction, reconstructing one view from the other might result in low quality reconstruction. Finally, CCA cannot benefit from additional non-parallel, single-view data. This puts it at a severe disadvantage in several real world situations, where in addition to some parallel two-view data, abundant single view data is available for one or both views.

Recently, Multimodal Autoencoders (MAEs) \citep{ngiam11} have been proposed to learn a common representation for two views/modalities. The idea in MAE is to train an autoencoder to perform two kinds of reconstruction. Given any one view, the model learns both self-reconstruction and cross-reconstruction (reconstruction of the other view). This makes the representations learnt to be predictive of each other. However, it should be noticed that the MAE does not
get any explicit learning signal encouraging it
to share the capacity of its common hidden layer between the views. In other words, it could develop units whose activation is dominated by a single view. This makes the MAE not suitable for transfer learning, since the views are not guaranteed to be projected to a common subspace. This is indeed verified by the results reported in \citep{ngiam11} where they show that CCA performs better than deep MAE for the task of transfer learning. 

These two approaches have complementary characteristics. On one hand, we have CCA and its variants which aim to produce correlated common representations but lack reconstruction capabilities. On the other hand, we have MAE which aims to do self-reconstruction and cross-reconstruction but does not guarantee correlated common representations. In this paper, we propose Correlational Neural Network (CorrNet) as a method for learning common representations which combines the advantages of the two approaches described above. The main characteristics of the proposed method can be summarized as follows:
\begin{itemize}
\item It allows for self/cross reconstruction. Thus, unlike CCA (and like MAE) it has predictive capabilities. This can be useful in applications where a missing view needs to be reconstructed from an existing view. 
\item Unlike MAE (and like CCA) the training objective used in CorrNet ensures that the common representations of the two views are correlated. This is particularly useful in applications where we need to match items from one view to their corresponding items in the other view.
\item CorrNet can be trained using Gradient Descent based optimization methods. Particularly, when dealing with large high dimensional data, one can use Stochastic Gradient Descent with mini-batches. Thus, unlike CCA (and like MAE) it is easy to scale CorrNet. 
\item The procedure used for training CorrNet can be easily modified to benefit from additional single view data. This makes CorrNet useful in many real world applications where additional single view data is available.
\end{itemize}

We evaluate CorrNet using four different experimental setups. First, we use the MNIST hand-written digit recognition dataset to compare CorrNet with other state of the art CRL approaches. In particular, we evaluate its (i) ability to self/cross reconstruct (ii) ability to produce correlated common representations and (iii) usefulness in transfer learning. In this setup, we use the left and right halves of the digit images as two views. Next, we use CorrNet for a transfer learning task where the two views of data come from two different languages. Specifically, we use CorrNet to project parallel documents in two languages to a common subspace. We then employ these common representations for the task of cross language document classification (transfer learning) and show that they perform better than the representations learned using other state of the art approaches. Third, we use CorrNet for the task of transliteration equivalence where the aim is to match a  name written using the script of one language (first view) to the same name written using the script of another language (second view). Here again, we demonstrate that with its ability to produce better correlated  common representations, CorrNet performs better than CCA and MAE. Finally, we employ CorrNet for a bigram similarity task and show that jointly learning words representations for two languages (two views) leads to better words representations. Specifically, representations learnt using CorrNet help to improve the performance of a bigram similarity task. We would like to emphasize that unlike other models which have been tested mostly in only one of these scenarios, we demonstrate the effectiveness of CorrNet in all these different scenarios. 

The remainder of this paper is organized as follows. In section \ref{sec:corrNet} we describe the architecture of CorrNet and outline a training procedure for learning its parameters. In section \ref{sec:deepCorrNet} we propose a deep variant of CorrNet. In section \ref{sec:relatedModels} we briefly discuss some related models for learning common representations. In section \ref{sec:analysisCorrNet} we present experiments to analyze the characteristics of CorrNet and compare it with CCA, KCCA and MAE. In section \ref{sec:deepCorrNetExperiments} we empirically compare Deep CorrNet with some other deep CRL methods. In sections \ref{sec:cldc}, \ref{sec:transEq}, and \ref{sec:bigram}, we report results obtained by using CorrNet for the tasks of cross language document classification, transliteration equivalence detection and bigram similarity respectively. Finally, we present concluding remarks in section \ref{sec:concFutureWork} and highlight possible future work.

\section{Correlational Neural Network}
\label{sec:corrNet}
As described earlier, our aim is to learn a common representation from two views of the same data such that: (i) any single view can be reconstructed from the common representation,  (ii) a single view can be predicted from the representation of another view and (iii) like CCA, the representations learned for the two views are correlated. The first goal above can be achieved by a conventional autoencoder. The first and second can be achieved together by a Multimodal autoencoder but it is not guaranteed to project the two views to a common subspace. We propose a variant of autoencoders which can work with two views of the data, while being explicitly trained to achieve all the above goals. In the following sub-sections, we describe our model and the training procedure.  

\subsection{Model}
We start by proposing a neural network architecture which contains three layers: an input layer, a hidden layer and an output layer. Just as in a conventional single view autoencoder, the input and output layers have the same number of units, whereas the hidden layer can have a different number of units.
For illustration, we consider a two-view input $\zz = (\xx,\yy)$. 
For all the discussions, $[\xx,\yy]$ denotes a concatenated vector of size $d_1 + d_2$.

\begin{figure}[ht]
\centering
\includegraphics[scale=0.6]{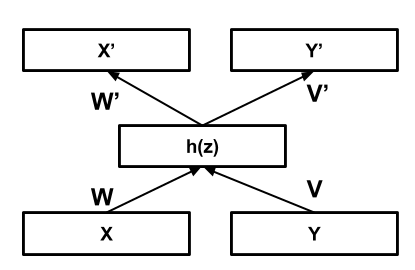}
\caption{Correlational Neural Network}
\label{fig:CNN}
\end{figure}

Given $\zz = (\xx,\yy)$, the hidden layer computes an encoded representation as follows:
\[ h(\zz) = f( \WW\xx + \VV\yy + \bb) \]
where $\WW$ is a $k \times d_1$ projection matrix, $\VV$ is a $k \times d_2$ projection matrix and $\bb$ is a $k\times 1$ bias vector. Function $f$ can be any non-linear activation function, for example \textit{sigmoid} or \textit{tanh}. The output layer then tries to reconstruct $\zz$ from this hidden representation by computing
\[ \zz' = g( [\WW'h(\zz), \VV'h(\zz)] + \bb')     \]
where $\WW'$ is a $d_1 \times k$ reconstruction matrix, $\VV'$ is a $d_2 \times k$ reconstruction matrix and $\bb'$ is a $(d_1 + d_2)\times 1$ output bias vector. Vector $\zz'$ is the reconstruction of $\zz$. Function $g$ can be any activation function. This architecture is illustrated in Figure~\ref{fig:CNN}. The parameters of the model are $\theta = \{ \WW, \VV, \WW', \VV', \bb, \bb' \}$. In the next sub-section we outline a procedure for learning these parameters.

\subsection{Training}
\label{subsec:training}
Restating our goals more formally, given a two-view data ${\cal Z} = \{(\zz_i)\}_{i=1}^N$ = $ \{(\xx_i,\yy_i)\}_{i=1}^N$, for each instance, $(\xx_i,\yy_i)$, we would like to:
\begin{itemize}
\item Minimize the self-reconstruction error, \textit{i.e.},  minimize the error in reconstructing $\xx_i$ from $\xx_i$ and $\yy_i$ from $\yy_i$.
\item Minimize the cross-reconstruction error, \textit{i.e.},  minimize the error in reconstructing $\xx_i$ from $\yy_i$ and $\yy_i$ from $\xx_i$.
\item Maximize the correlation between the hidden representations of both views.
\end{itemize}
We achieved this by finding the parameters $\theta = \{ \WW, \VV, \WW', \VV', \bb, \bb' \}$ which minimize the following objective function:
\[  \mathcal{J}_{\cal Z}(\theta) = \sum_{i=1}^N ( L(\zz_i,g(h(\zz_i))) + L(\zz_i,g(h(\xx_i))) + L(\zz_i,g(h(\yy_i))) ) - \lambda~{\rm corr}(h(X),h(Y))       \]

\[ {\rm corr}(h(X),h(Y)) =  \frac{\sum_{i=1}^N (h(\xx_i) - \overline{h(X)})(h(\yy_i) - \overline{h(Y)})  }{\sqrt{\sum_{i=1}^N (h(\xx_i) - \overline{h(X)})^2 \sum_{i=1}^N (h(\yy_i) - \overline{h(Y)})^2    }} 
          \]
where $L$ is the reconstruction error, $\lambda$ is the scaling parameter to scale the fourth term with respect to the remaining three terms, $\overline{h(X)}$ is the mean vector for the hidden representations of the first view and $\overline{h(Y)}$ is the mean vector for the hidden representations of the second view.  If all dimensions in the input data take binary values then we use cross-entropy as the reconstruction error otherwise we use squared error loss as the reconstruction error. For simplicity, we use the shorthands $h(\xx_i)$ and $h(\yy_i)$ to note the representations $h((\xx_i,0))$ and $h((0,\yy_i))$ that are based only on a single view\footnote{They represent the generic functions $h_X$ and $h_Y$ mentioned in the introduction. }. For each data point with 2 views $x$ and $y$, $h(\xx_i)$ just means that we are computing the hidden representation using only the $x$-view. Or in other words, in equation h(\zz) = f( \WW\xx + \VV\yy + \bb), we set $\yy$=0. So, h(\xx) = f( \WW\xx + \bb). $h(\xx)$ = $h(\xx,0)$ is not a choice per se, but a notation we are defining. $h(\zz)$, $h(\xx)$ and $h(\yy)$ are certainly not guarantied to be identical, though training will gain in making them that way, because of the various reconstruction terms. The correlation term in the objective function is calculated considering the hidden representation as a random vector.

In words, the objective function decomposes as follows. The first term is the usual autoencoder objective function which helps in learning meaningful hidden representations. The second term ensures that both views can be predicted from the shared representation of the first view alone. The third term ensures that both views can be predicted from the shared representation of the second view alone.  The fourth term interacts with the other objectives to make sure that the hidden representations are highly correlated, so as to encourage the hidden units of the representation to be shared between views. 

We can use stochastic gradient descent (SGD) to find the optimal parameters. For all our experiments, we used mini-batch SGD. The fourth term in the objective function is then approximated based on the statistics of a minibatch. Approximating second order statistics using minibatches for training was also used successfully in the batch normalization training method of~\citet{IoffeS2015}.

The model has four hyperparameters: (i) the number of units in its hidden layer, (ii) $\lambda$, (iii) mini-batch size, and (iv) the SGD learning rate. The first hyperparameter is dependent on the specific task at hand and can be tuned using a validation set (exactly as is done by other competing algorithms). The second hyperparameter is only to ensure that the correlation term in the objective function has the same range as the reconstruction errors. This is again easy to approximate based on the given data. The third hyperparameter approximates the correlation of the entire dataset and larger mini-batches are preferred over smaller mini-batches. The final hyperparameter, the learning rate is common for all neural network based approaches.

Once the parameters are learned, we can use the CorrNet to compute representations of views that can potentially generalize across views. Specifically, given a new data instance for which only one view is available, we can compute its corresponding representation ($h(\xx)$ if $\xx$ is observed or $h(\yy)$ if $\yy$ is observed) and use it as the new data representation.

\subsection{Using additional single view data}
\label{subsec:singleViewCorrNet}
In practice, it is often the case that we have abundant single view data and comparatively little two-view data. For example, in the context of text documents from two languages  ($X$ and $Y$), typically the amount of monolingual (single view) data available in each language is much larger than parallel (two-view) data available between $X$ and $Y$. Given the abundance of such single view data, it is desirable to exploit it in order to improve the learned representation. CorrNet can achieve this, by using the single view data to improve the self-reconstruction error as explained below. 

Consider the case where, in addition to the data ${\cal Z} = \{(\zz_i)\}_{i=1}^N$ = $ \{(\xx_i,\yy_i)\}_{i=1}^N$, we also have access to the single view data ${\cal X} = \{(\xx_i)\}_{i=N+1}^{N_1}$ and ${\cal Y} = \{(\yy_i)\}_{i=N+1}^{N_2}$. Now, during training, in addition to using ${\cal Z}$ as explained before, we also use ${\cal X}$ and ${\cal Y}$ by suitably modifying the objective function so that it matches that of a  conventional autoencoder. Specifically, when we have only $\xx_i$, then we could try to minimize
\[  \mathcal{J}_{\cal X}(\theta) = \sum_{i=N+1}^{N_1}  L(\xx_i,g(h(\xx_i)))  \]
and similarly for $\yy_i$. 

In all our experiments, when we have access to all three types of data (\textit{i.e.}, ${\cal X}$, ${\cal Y}$ and ${\cal Z}$), we construct 3 sets of mini-batches by sampling data from ${\cal X}$, ${\cal Y}$ and ${\cal Z}$ respectively. We then feed these mini-batches in random order to the model and perform a gradient update based on the corresponding objective function.

\section{Deep Correlational Neural Networks}
\label{sec:deepCorrNet}
An obvious extension for CorrNets is to allow for multiple hidden layers. The main motivation for having such Deep Correlational Neural Networks is that a better correlation between the views of the data might be achievable by more non-linear representations. 

\begin{figure}[ht]
\centering
\includegraphics[scale=0.3]{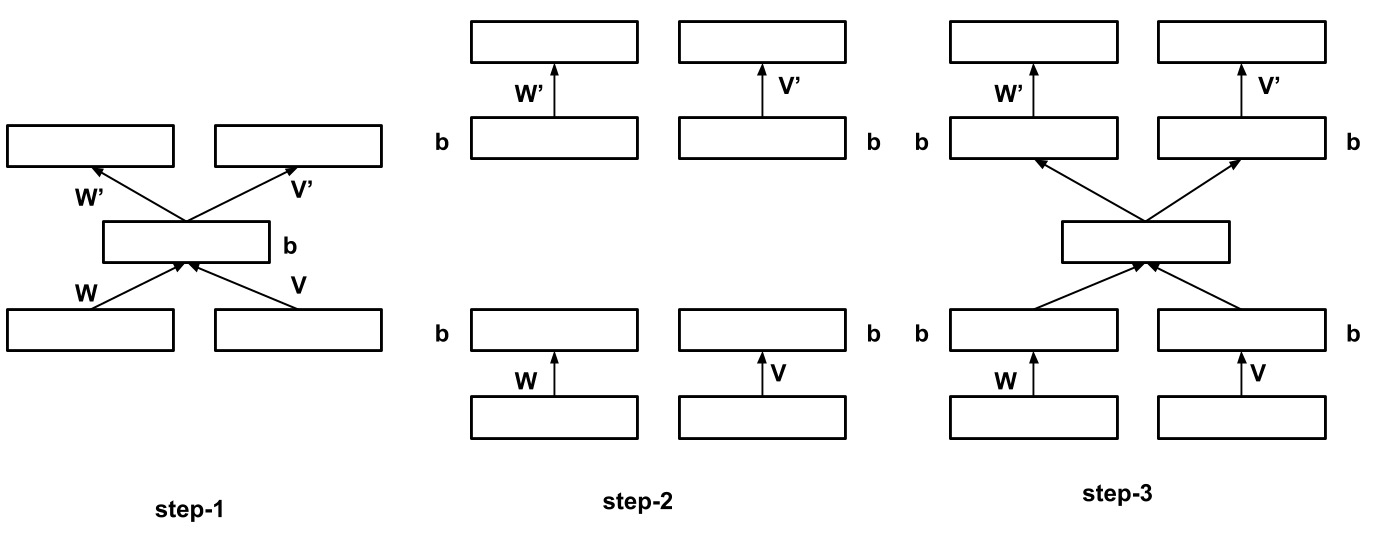}
\caption{Stacking CorrNet to create Deep Correlartional Neural Network.}
\label{DCNN}
\end{figure}

We use the following procedure to train a Deep CorrNet.
\begin{enumerate}
\item Train a shallow CorrNet with the given data  (see step-1 in Figure~\ref{DCNN}). At the end of this step, we have learned the parameters $\WW$, $\VV$ and $\bb$.
\item Modify the CorrNet model such that the first input view connects to a hidden layer using weights $\WW$ and bias $\bb$. Similarly connect the second view to a hidden layer using weights $\VV$ and bias $\bb$. We have now decoupled the common hidden layer for each view (see step-2 in Figure \ref{DCNN}). 
\item Add a new common hidden layer which takes its input from the hidden layers created at step 2. We now have a CorrNet which is one layer deeper (see step-3 in Figure \ref{DCNN}).
\item Train the new Deep CorrNet on the same data.
\item Repeat steps 2, 3 and 4, for as many hidden layers as required.
\end{enumerate}

We would like to point out that we could have followed the procedure described in \cite{aps2} for training Deep CorrNet. In \cite{aps2}, they learn deep representation for each view separately and use it along with a shallow CorrNet to learn a common representation. However, feeding non-linear deep representations to a shallow CorrNet makes it harder to train the CorrNet. Also, we chose not to use the deep training procedure described in \cite{ngiam11} since the objective function used by them during pre-training and training is different. Specifically, during pre-training the objective is to minimize self reconstruction error whereas during training the objective is to minimize both self and cross reconstruction error. In contrast, in the stacking procedure outlined above, the objectives during training and pre-training are aligned. 

Our current training procedure for deep CorrNet is similar to greedy layerwise pretraining of deep autoencoders. We believe that this procedure is more faithful to the global training objective of Corrnet and it works well. We do not have strong empirical evidence that this is superior to other methods such as the one described in \cite{aps2} and \cite{ngiam11}. When we have less parallel data, using method described in \cite{aps2} makes more sense and each method has its own advantages. We leave a detailed comparison of these different alternatives of Deep CorrNet as future work.

\section{Related Models}
\label{sec:relatedModels}

In this section, we describe other related models for common representation learning. We restrict our discussion to CCA based methods and Neural Network based methods only.

Canonical Correlation Analysis (CCA) \citep{cca} and its variants, such as regularized CCA \citep{rcca2, rcca3, rcca1} are the de-facto approaches used for learning common representation for two different views in the literature \citep{Udupa:10,cca_app}. Kernel CCA \citep{kcca, hardoon} which is another variant of CCA uses the standard kernel trick to find pairs of non-linear projections of the two views. Deep CCA, a deep version of CCA is also introduced in \citep{dcca}. One issue with CCA is that it is not easily scalable. Even though there are several works on scaling CCA (see \citep{fcca}), they are all approximations to CCA and hence lead to a decrease in the performance. Also is not very trivial to extend CCA to multiple views. However there are some recent work along this line \citep{mcca,tcca} which require complex computations. Lastly, conventional CCA based models can work only with parallel data. However, in real life situations, parallel data is costly when compared to single view data. The inability of CCA to leverage such single view data acts as a drawback in many real world applications. Representation Constrained CCA (RCCCA) \citep{rccca} is one such model which can benefit from both single view data and multiview data. It effectively uses a weighted combination of PCA (for single view) and CCA (for two views) by minimizing self-reconstruction errors and maximizing correlation. CorrNet, in contrast,  minimizes both self and cross reconstruction error while maximizing correlation. RCCCA can also be considered as a linear version of DCCAE proposed in \cite{wang}.

\citet{nlcca} is one of the earliest Neural Network based model for nonlinear CCA. This method uses three feedforward neural networks. The first neural network is a double-barreled architecture where two networks project the views to a single unit such that the projections are maximally correlated. This network is first trained to maximize the correlation. Then the inverse mapping for each view is learnt from the corresponding canonical covariate representation by minimizing the reconstruction error. There are clear differences between this Neural CCA model and CorrNet. First, CorrNet is a single neural network which is trained with a single objective function while Neural CCA has three networks trained with different objective functions. Second, Neural CCA does only correlation maximization and self-reconstruction, whereas CorrNet does correlation maximization, self-reconstruction and cross-reconstruction, all at the same time. 

Multimodal Autoencoder (MAE) \citep{ngiam11} is another Neural Network based CRL approach. Even though the architecture of MAE is similar to that of CorrNet there are clear differences in the training procedure used by the two. Firstly, MAE only aims to minimize the following three errors: (i) error in reconstructing $z_i$ from $x_i$ ($E_1$),  (ii) error in reconstructing $z_i$ from $y_i$ ($E_2$) and (iii) error in reconstructing $z_i$ from $z_i$ ($E_3$). More specifically, unlike the fourth term in our objective function, the objective function used by MAE does not contain any term which forces the network to learn correlated common representations. Secondly, there is a difference in the manner in which these terms are considered during training. Unlike CorrNet, MAE only considers one of the above terms at a time. In other words, given an instance $z_i = (x_i, y_i)$ it first tries to minimize $E_1$ and updates the parameters accordingly. It then tries to minimize $E_2$ followed by $E_3$. Empirically, we observed that a training procedure which considers all three loss terms together performs better than the one which considers them separately (Refer Section 5.5). 

Deep Canonical Correlation Analysis (DCCA) \citep{dcca} is a recently proposed Neural Network approach for CCA. DCCA employs two deep networks, one per view. The model is trained in such a way that the final layer projections of the data in both the views are maximally correlated. DCCA maximizes only correlation whereas CorrNet maximizes both, correlation and reconstruction ability. Deep Canonically Correlated Auto Encoders (DCCAE) \citep{wang} (developed in parallel with our work) is an extension of DCCA which considers self reconstruction and correlation. Unlike CorrNet it does not consider cross-reconstruction.

\section{Analysis of Correlational Neural Networks}
\label{sec:analysisCorrNet}
In this section, we perform a set of experiments to compare  CorrNet, CCA \citep{cca}, Kernel CCA (KCCA) \citep{kcca} and MAE \citep{ngiam11} based on:   

\begin{itemize}
\item ability to reconstruct a view from itself
\item ability to reconstruct one view given the other 
\item ability to learn correlated common representations for the two views
\item usefulness of the learned common representations in transfer learning.
\end{itemize}       
    For CCA, we used a C++ library called \textit{dlib} \citep{dlib09}. For KCCA, we used an implementation provided by the authors of \citep{kccacode}. We implemented CorrNet and MAE using Theano \citep{theano}.
    
\subsection{Data Description}
\label{subsec:data_description}
We used the standard MNIST handwritten digits image dataset for all our experiments. This data consists of 60,000 train images and 10,000 test images. Each image is a 28 * 28 matrix of pixels; each pixel representing one of 256 grayscale values. We treated the left half of the image as one view and the right half of the image as another image. Thus each view contains $14*28 = 392$ dimensions. We split the train images into two sets. The first set contains 50,000 images and is used for training. The second set contains 10,000 images and is used as a validation set for tuning the hyper-parameters of the four models described above. 

\subsection{Performance of Self and Cross Reconstruction}
Among the four models listed above, only CorrNets and MAE have been explicitly trained to construct a view from itself as well as from the other view. So, in this sub-section, we consider only these two models. Table \ref{trecon} shows the Mean Squared Errors (MSEs) for self and cross reconstruction when the left half of the image is used as input.  

\begin{table}[H]
\centering
\begin{tabular}{ |c| c| c| }
\hline
 \textbf{Model}& \textbf{MSE for self reconstruction} & \textbf{MSE for cross reconstruction}   \\
 \hline
 CorrNet & 3.6 &	4.3   \\
  
 MAE &  \textbf{2.1} & \textbf{4.2}   \\
 \hline
 \end{tabular} 
\caption{Mean Squared Error for CorrNet and MAE for self reconstruction and cross reconstruction}

\label{trecon}
\end{table}

The above table suggests that CorrNet has a higher self reconstruction error and almost the same cross reconstruction error as that of MAE. This is because unlike MAE, in CorrNet, the emphasis is on maximizing the correlation between the common representations of the two views. This goal captured by the fourth term in the objective function obviously interferes with the goal of self reconstruction. As we will see in the next sub-section, the embeddings learnt by CorrNet for the two views are better correlated even though the self-reconstruction error is sacrificed in the process.

\begin{figure}[h]
\centering
\includegraphics{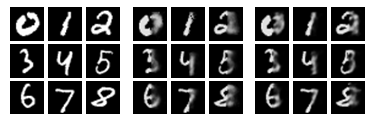}
\caption{Reconstruction of right half of the image given the left half. First block shows the original images, second block shows images where the right half is reconstructed by CorrNet and the third block shows images where the right half is reconstructed by MAE.}
\label{fig5}
\end{figure}

Figure \ref{fig5} shows the reconstruction of the right half from the left half for a few sample images. The figure reiterates our point that both CorrNet and MAE are equally good at cross reconstruction.

\subsection{Correlation between representations of two views}
As mentioned above, in CorrNet we emphasize on learning highly correlated representations for the two views. To show that this is indeed the case, we follow \citep{dcca} and calculate the total/sum correlation captured in the 50 dimensions of the common representations learnt by the four models described above. The training, validation and test sets used for this experiment were as described in section \ref{subsec:data_description}. The results are reported in Table \ref{tsum}. 

\begin{table}[H]
\centering
\begin{tabular}{ |c| c|  }
\hline
 \textbf{Model}& \textbf{Sum Correlation}   \\
 \hline
 CCA & 17.05  \\
 KCCA &  30.58 \\
 MAE & 24.40  \\
 CorrNet &  \textbf{45.47}  \\
  \hline
 \end{tabular} 
\\
\caption{Sum/Total correlation captured in the 50 dimensions of the common representations learned by different models using MNIST data.}
\label{tsum}
\end{table}

The total correlation captured in the 50 dimensions learnt by CorrNet is clearly better than that of the other models. 

\begin{figure}[H]
\centering
\includegraphics[scale=0.7]{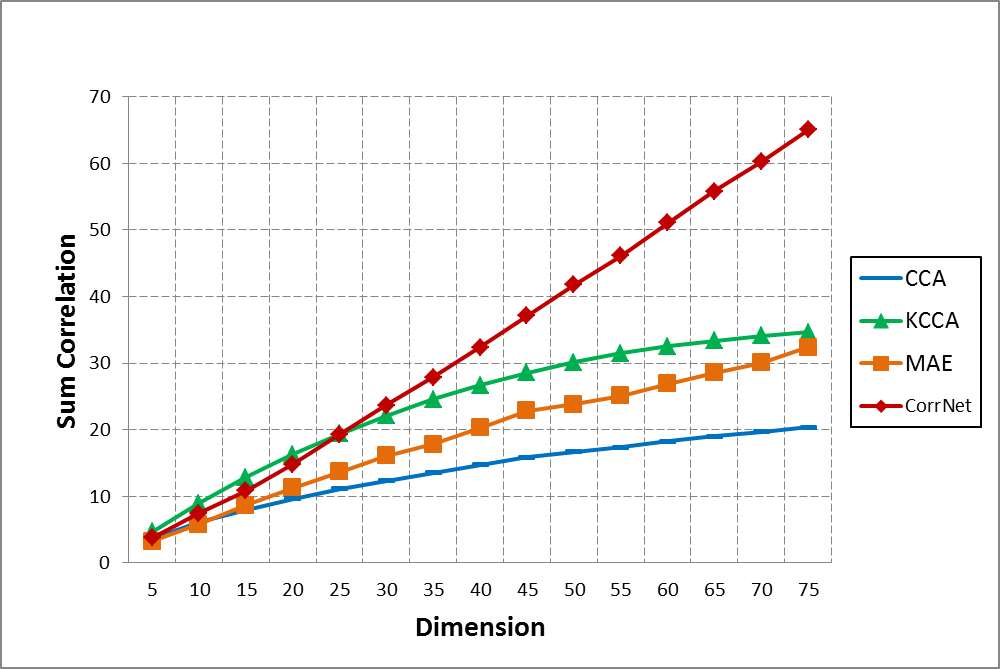}
\caption{Sum/Total correlation as a function of the number of dimensions in the common representations learned by different models using MNIST data.}
\label{fig4}
\end{figure}

Next, we check whether this is indeed the case when we change the number of dimensions. For this, we varied the number of dimensions from 5 to 80 and plotted the sum correlation for each model (see Figure \ref{fig4}). For all the models, we tuned the hyper-parameters for $dim=50$ and used the same hyper-parameters for all dimensions.

Again, we see that CorrNet clearly outperforms the other models. CorrNet thus achieves its primary goal of producing correlated embeddings with the aim of assisting transfer learning. 

\subsection{Transfer Learning across views}
\label{transexp}
To demonstrate transfer learning, we take the task of predicting digits from only one half of the image. We first learn a common representation for the two views using 50,000 images from the MNIST training data. For each training instance, we take only one half of the image and compute its 50 dimensional common representation using one of the models described above. We then train a classifier using this representation. For each test instance, we consider only the other half of the image and compute its common representation. We then feed this representation to the classifier for prediction. We use the linear SVM implementation provided by \citep{sklearn} as the classifier for all our experiments. For all the models considered in this experiment, representation learning is done using 50,000 train images and the best hyperparameters are chosen using the 10,000 images from the validation set. With the chosen model, we report 5-fold cross validation accuracy using 10,000 images available in the standard test set of MNIST data. We report accuracy for two settings (i) Left to Right (training on left view, testing on right view) and (ii) Right to Left (training on right view, testing on left view).

\begin{table}[H]
\centering
\begin{tabular}{ |c| c| c| }
\hline

 \textbf{Model}& \textbf{Left to Right} & \textbf{Right to Left}   \\
 \hline
 CCA & 65.73 & 65.44  \\
 KCCA & 68.1 & 75.71 \\
 MAE & 64.14 & 68.88  \\
 CorrNet &  \textbf{77.05} & \textbf{78.81}  \\
 \hline
 Single view & 81.62 & 80.06\\
 \hline
 \end{tabular} 
\\
\caption{Transfer learning accuracy using the representations learned using different models on the MNIST dataset.}
\end{table}

Single view corresponds to the classifier trained and tested on same view. This is the upper bound for the performance of any transfer learning algorithm. Once again, we see that CorrNet performs significantly better than the other models. To verify that this holds even when we decrease the data for learning common representation to 10000 images. The results as reported in Table \ref{t10} show that even with less data, CorrNet perform betters than other models.

\begin{table}[H]
\centering
\begin{tabular}{ |c| c| c| }
\hline
 \textbf{Model}& \textbf{Left to Right} & \textbf{Right to Left}   \\
 \hline
 CCA & 66.13 & 66.71  \\
 KCCA & 70.68  &  70.83 \\
 MAE & 68.69 & 72.54  \\
 CorrNet &  \textbf{76.6} & \textbf{79.51}  \\
\hline
Single view & 81.62 & 80.06 \\
 
 \hline
 \end{tabular} 
\\
\caption{Transfer learning accuracy using the representations learned using different models trained with 10000 instances from the MNIST dataset.}
\label{t10}
\end{table}

\subsection{Relation with MAE}
At the face of it, it may seem that both CorrNet and MAE differ only in their objective functions. Specifically, if we remove the last correlation term from the objective function of CorrNet then it would become equivalent to MAE. To verify this, we conducted experiments using both MAE and CorrNet without the last term (say CorrNet(123)). When using SGD to train the networks, we found that the performance is almost similar. However, when we use some advanced optimization technique like RMSProp, CorrNet(123) starts performing better than MAE. The results are reported in Table \ref{tcomp}.

\begin{table}[H]
\centering
\begin{tabular}{ |c| c| c| c| }
\hline
 \textbf{Model}& \textbf{Optimization} & \textbf{Left to Right} & \textbf{Right to Left}   \\
 \hline
 MAE & SGD & 63.9 & 67.98  \\
 CorrNet(123) & SGD & 63.89 & 67.93  \\
 MAE & RMSProp & 64.14 & 68.88  \\
 CorrNet(123) & RMSProp & 67.82 & 72.13  \\
 
 \hline
 \end{tabular} 
\\
\caption{Results for transfer learning across views}
\label{tcomp}
\end{table}

This experiment sheds some light on why CorrNet is better than MAE. Even though the objective of MAE and CorrNet(123) is same, MAE tries to solve it in a stochastic way which adds more noise. However, CorrNet(123) performs better since it is actually working on the combined objective function and not the stochastic version (one term at a time) of it.

\subsection{Analysis of Loss Terms}
The objective function defined in Section \ref{subsec:training} has the following four terms:
\begin{itemize}
\item $L_1 = \sum_{i=1}^N  L(\zz_i,g(h(\zz_i))$
\item $L_2 = \sum_{i=1}^N  L(\zz_i,g(h(\xx_i))$
\item $L_3 = \sum_{i=1}^N  L(\zz_i,g(h(\yy_i))$
\item $L_4 = \lambda ~{\rm corr}(h(X),h(Y))$
\end{itemize}
In this section, we analyze the importance of each of these terms in the loss function. For this, during training, we consider different loss functions which contain different combinations of these terms. In addition, we consider four more loss terms for our analysis. 
\begin{itemize}
\item $L_5 = \sum_{i=1}^N  L(\yy_i,g(h(\xx_i)) $
\item $L_6 = \sum_{i=1}^N  L(\xx_i,g(h(\yy_i)) $
\item $L_7 = \sum_{i=1}^N  L(\xx_i,g(h(\xx_i)) $
\item $L_8 = \sum_{i=1}^N  L(\yy_i,g(h(\yy_i)) $
\end{itemize}
where $L5$ and $L6$ essentially capture the loss in reconstructing only one view (say, $\xx_i$) from the other view ($\yy_i$) while $L7$ and $L8$ capture the loss in self reconstruction.  

For this, we first learn common representations using different loss functions as listed in the first column of Table \ref{tloss}. We then repeated the transfer learning experiments using common representations learned from each of these models. For example, the sixth row in the table shows the results when the following loss function is used for learning the common representations. 
\[  \mathcal{J}_{\cal Z}(\theta) = L_1 + L_2 + L_3 + L_4 \]
which is the same as that used in CorrNet.

\begin{table}[H]
\centering
\begin{tabular}{ |c| c| c| }
\hline
 \textbf{Loss function used for training}&  \textbf{Left to Right} & \textbf{Right to Left}   \\
 \hline
 $L_1$ &  24.59 & 22.56 \\
 $L_1 + L_4$ &  65.9 & 67.54  \\
 $L_2+L_3$ &    71.54 & 75 \\
 $L_2+L_3+L_4$ &   76.54 & \textbf{80.57}  \\
 $L_1+L_2+L_3$ &   67.82 & 72.13  \\
 $L_1+L_2+L_3+L_4$ &   \textbf{77.05} & 78.81  \\
 $L_5+L_6$ &   35.62 & 32.26 \\
 $L_5+L_6+L_4$  & 62.05 & 63.49 \\
 $L_7+L_8$ & 10.26   & 10.33 \\
 $L_7+L_8+L_4$  & 73.03 &   76.08 \\
 \hline
 \end{tabular} 
\\
\caption{Comparison of the performance of transfer learning with representations learned using different loss functions.}
\label{tloss}
\end{table}

Each even numbered row in the table reports the performance when the correlation term ($L_4$) was used in addition to the other terms in the row immediately before it. A pair-wise comparison of the numbers in each even numbered row with the row immediately above it suggests that the correlation term ($L_4$) in the loss function clearly produces representations which lead to better transfer learning.

\section{Experiments using Deep Correlational Neural Network}
\label{sec:deepCorrNetExperiments}

In this section, we evaluate the performance of the deep extension of CorrNet. Having already compared with MAE in the previous section, we focus our evaluation here on a comparison with DCCA~\citep{dcca}. All the models were trained using 10000 images from the MNIST training dataset and we computed the sum correlation and transfer learning accuracy for each of these models. For transfer learning, we use the linear SVM implementation provided by \citep{sklearn} for all our experiments and do 5-fold cross validation using 10000 images from MNIST test data. We report results for two settings (i) Left to Right (training on left view, testing on right view) and (ii) Right to Left (training on right view, testing on left view). These results are summarized in Table \ref{tdeep}. In this Table, model-$x$-$y$ means a model with $x$ units in the first hidden layer and $y$ units in second hidden layer. For example, CorrNet-500-300-50 is a Deep CorrNet with three hidden layers containing 500, 300 and 50 units respectively. The third layer containing 50 units is used as the common representation.

\begin{table}[H]
\centering
\begin{tabular}{ |c| c| c| c| }
\hline
 \textbf{Model}& \textbf{Sum Correlation} & \textbf{Left to Right} & \textbf{Right to Left}   \\
 \hline
  CorrNet-500-50 &  \textbf{47.21} & 77.68  & 77.95  \\
 DCCA-500-50 & 33.00  & 66.41 & 64.65  \\
 \hline
  CorrNet-500-300-50 & 45.634 & \textbf{80.46} & \textbf{80.47}  \\
   DCCA-500-500-50 & 33.77  & 70.06 & 72.43  \\
 \hline
 \end{tabular} 
\\
\caption{Comparison of sum correlation and transfer learning performance of different deep models}
\label{tdeep}
\end{table}

Both the Deep CorrNets (CorrNet-500-50 and CorrNet-500-300-50) clearly perform better than the corresponding DCCA. We notice that for both the transfer learning tasks, the 3-layered CorrNet (CorrNet-500-300-50) performs better than the 2-layered CorrNet (CorrNet-500-50) but the sum correlation of the 2-layered CorrNet is better than that of the 3-layered CorrNet.

\section{Cross Language Document Classification}
\label{sec:cldc}
In this section, we will learn bilingual word representations using CorrNet and use these representations for the task of cross language document classification. We experiment with three language pairs and show that our approach achieves state-of-the-art performance.

Before we discuss bilingual word representations let us consider the task of learning word representations for a single language. Consider a language $X$ containing $d$ words in its vocabulary. We represent a sentence in this language using a binary bag-of-words representation $\xx$. Specifically, each dimension $x_i$ is set to 1 if the $i^{\rm th}$ vocabulary word is present in the sentence, 0 otherwise.  We wish to learn a $k$-dimensional vectorial
representation of each word in the vocabulary from a training set of sentence
bags-of-words $\{\xx_i\}_{i=1}^N$.

We propose to achieve this by using a CorrNet which works with only a single view of the data (see section \ref{subsec:singleViewCorrNet}). Effectively, one can view a CorrNet as encoding
an input bag-of-words $\xx$ as the sum of the columns in $\WW$ corresponding to the words that are present in $\xx$, followed by a non-linearity. Thus, we can view $\WW$ as a
matrix whose columns act as
vector representations (embeddings) for each word. 

Let's now assume that for each sentence bag-of-words $\xx$ in some
source language $X$, we have an associated bag-of-words $\yy$ for this sentence translated in some target language $Y$ by a human expert. 
Assuming we have a training set of such $(\xx,\yy)$ pairs, we'd like to learn representations in both languages that are aligned,
such that pairs of translated words have similar representations. The CorrNet can allow us to achieve this. Indeed, it will effectively learn word representations (the columns of $\WW$ and $\VV$) that are not only informative about the words present in sentences of each language, but will also ensure that the representations' space is aligned between language, as required by the cross-view reconstruction terms and the correlation term.

Note that, since the binary bags-of-words are very high-dimensional (the dimensionality corresponds to the
size of the vocabulary, which is typically large), reconstructing each binary bag-of-word will be slow. Since we will later be training on millions of sentences, training on each individual sentence bag-of-words will be expensive. Thus, we propose a simple trick, which exploits the bag-of-words structure of the input. Assuming we are performing mini-batch training (where a mini-batch contains a list of the bags-of-words of adjacent sentences), we simply propose to merge the bags-of-words of the mini-batch into a single bag-of-words and perform an update based on that merged bag-of-words. The resulting effect is that each update is as efficient as in stochastic gradient descent, but the number of updates per training epoch is divided by the mini-batch size.
As we'll see in the experimental section, this trick produces good word representations, while sufficiently reducing training time.
We note that, additionally, we could have used the stochastic approach proposed by \citet{DauphinY2011} for reconstructing binary bag-of-words representations of documents, to further improve the efficiency of training. They use importance sampling to avoid reconstructing the whole $V$-dimensional input vector.

\subsection{Document representations}
\label{sec:docrep}
Once we learn the language specific word representation
matrices $\WW$ and $\VV$ as described above, we can use them to construct document representations, by using their columns as word vector representations. Given a document ${\bf d}$, we represent it as the tf-idf weighted sum of its words' representations: $\psi_X({\bf d}) = \WW \textit{tf-idf}({\bf d})$ for language $X$ and $\psi_Y({\bf d}) = \VV \textit{tf-idf}({\bf d})$ for language $Y$, where $\textit{tf-idf}({\bf d})$ is the tf-idf weight vector of document ${\bf d}$.

We use the document representations thus obtained to train our document classifiers, in the cross-lingual document classification task described in Section~\ref{exp}.

\subsection{Related Work on Multilingual Word Representations}
\label{sec:related_work}

Recent work that has considered the problem of learning bilingual representations of words usually has relied on word-level
alignments. \citet{KlementievA2012} propose to train simultaneously two neural
network languages models, along with a regularization term that
encourages pairs of frequently aligned words to have similar word
embeddings. Thus, the use of this regularization term requires to first obtain
word-level alignments from parallel corpora. 
\citet{ZhouW2013} use a similar approach, with a different form
for the regularizer and neural network language models as
in~\citep{CollobertR2011}. In our work, we specifically investigate whether
a method that does not rely on word-level alignments can
learn comparably useful multilingual embeddings in the
context of document classification.

Looking more generally at neural networks that learn multilingual
representations of words or phrases, we mention the work of \citet{GaoJ2014}
which showed that a useful linear mapping between {\it separately
  trained} monolingual skip-gram language models could be learned. 
They too however rely on the specification of pairs of words
in the two languages to align. \citet{MikolovT2013} also propose a method
for training a neural network to learn useful representations
of phrases, in the context of
a phrase-based translation model. In this case, phrase-level
alignments (usually extracted from word-level alignments)
are required. Recently, \cite{HermannK2014,Hermann:2014:ICLR},  proposed neural network architectures
and a margin-based training objective that, as in this work, does not
rely on word alignments. We will briefly discuss this work in the experiments section. A tree based bilingual autoencoder with similar objective function is also proposed in \citep{aps}. 

\subsection{Experiments}
\label{exp}

The technique proposed in this work enable us to learn bilingual embeddings which capture cross-language similarity between words. We propose to evaluate the quality of these embeddings by using them for the task of cross-language document classification. We followed closely the setup used by \citet{KlementievA2012} and compare with their method, for which word representations are publicly available\footnote{\url{http://klementiev.org/data/distrib/}}. The set up is as follows. A labeled data set of documents in some language $X$ is available to train a classifier, however we are interested in classifying documents in a different language $Y$ at test time. To achieve this, we leverage some bilingual corpora, which is not labeled with any document-level categories. This bilingual corpora is used to learn document representations that are coherent between languages $X$ and $Y$. The hope is thus that we can successfully apply the classifier trained on document representations for language $X$ directly to the document representations for language $Y$. Following this setup, we performed experiments on 3 data sets of language pairs: English/German (EN/DE), English/French (EN/FR) and English/Spanish (EN/ES).

For learning the bilingual embeddings, we used sections of the Europarl corpus~\citep{europ} which contains roughly 2 million parallel sentences. We considered 3 language pairs. We used the same pre-processing as used by \citet{KlementievA2012}. We tokenized the sentences using NLTK \citep{nltk}, removed punctuations and lowercased all words. We did not remove stopwords.   

As for the labeled document classification data sets, they were extracted from sections of the Reuters RCV1/RCV2 corpora, again for the 3 pairs considered in our experiments.
Following \citet{KlementievA2012}, we consider only documents which were assigned exactly one of the 4 top level categories in the topic hierarchy (CCAT, ECAT, GCAT and MCAT).
These documents are also pre-processed using a similar procedure as that used for the Europarl corpus. We used the same vocabularies as those used by \citet{KlementievA2012} (varying in size between $35,000$ and $50,000$).

Models were trained for up to 20 epochs using the same data as described earlier. We used mini-batch (of size 20) stochastic gradient descent. All results are for word embeddings of size $D=40$, as in \citet{KlementievA2012}. Further, to speed up the training for CorrNet
we merged each 5 adjacent sentence pairs into a single training instance, as described earlier. 
For all language pairs, $\lambda$ was set to $4$. The other hyperparameters were tuned to each task using a training/validation set split of 80\% and 20\% and using the performance on the validation set of an averaged perceptron trained on the smaller training set portion (notice that this corresponds to a monolingual classification experiment, since the general assumption is that no labeled data is available in the test set language).

We compare our models with the following approaches:
\begin{itemize}
\item \cite{KlementievA2012}: This model uses word embeddings learned by a multitask neural network language model with a regularization term that
encourages pairs of frequently aligned words to have similar word
embeddings. From these embeddings, document representations are computed as described in Section~\ref{sec:docrep}. 
\item MT: Here, test documents are translated to the language of the training documents using a standard phrase-based MT system,  MOSES\footnote{\url{http://www.statmt.org/moses/}} which was trained using default parameters and a 5-gram language model on the Europarl corpus (same as the one used for inducing our bilingual embeddings).

\item Majority Class: Test documents are simply assigned the most frequent class in the training set.
\end{itemize}

For the EN/DE language pairs, we directly report the results from \citet{KlementievA2012}. For the other pairs (not reported in  \citet{KlementievA2012}), we used the embeddings available online and performed the classification experiment ourselves. Similarly, we generated the MT baseline ourselves. 

Table \ref{tab:title} summarizes the results. They were obtained using 1000 RCV training examples. We report results in both directions, \textit{i.e.}\ language $X$ to $Y$ and vice versa. The best performing method in all the pairs except one is CorrNet. In particular, CorrNet often outperforms the approach of \citet{KlementievA2012} by a large margin. 

In the last row of the table, we also include the results of some recent work by \cite{HermannK2014,Hermann:2014:ICLR}. They proposed two neural network architectures for learning word and document representations using sentence-aligned data only. Instead of an autoencoder paradigm, they propose a margin-based objective that aims to make the representation of aligned sentences closer than non-aligned sentences. While their trained embeddings are not publicly available, they report results for the EN/DE classification experiments, with representations of the same size as here ($D=40$) and trained on 500K EN/DE sentence pairs. Their best model in that setting reaches accuracies of 83.7\% and 71.4\% respectively for the EN~$\rightarrow$~DE and DE~$\rightarrow$~EN tasks. One clear advantage of our model is that unlike their model, it can use additional monolingual data. Indeed, when we train CorrNet with 500k EN/DE sentence pairs, plus monolingual RCV documents (which come at no additional cost), we get accuracies of 87.9\% (EN~$\rightarrow$~DE) and 76.7\% (DE~$\rightarrow$~EN), still improving on their best model. If we do not use the monolingual data, CorrNet's performance is worse but still competitive at 86.1\% for EN~$\rightarrow$~DE and 68.8\% for DE~$\rightarrow$~EN. Finally, without constraining $D$ to $40$ (they use $128$) and by using additional French data, the best results of \citet{HermannK2014} are 88.1\% (EN~$\rightarrow$~DE) and 79.1\% (DE~$\rightarrow$~EN), the later being, to our knowledge, the current state-of-the-art (as reported in the last row of Table \ref{tab:title})\footnote{After we published our results in \citep{aps}, \citet{soyer2015leveraging} have improved the performance for EN$\rightarrow$DE and DE$\rightarrow$EN to 92.7\% and 82.4\% respectively.}.

\begin {table}[H]
\begin{center}
\begin{footnotesize}
\begin{tabular}{|l|l|l|l|l|l|l|}
\hline
{\bf } & EN $\rightarrow$ DE & DE $\rightarrow$ EN & EN $\rightarrow$ FR & FR $\rightarrow$ EN & EN $\rightarrow$ ES & ES $\rightarrow$ EN \\
\hline\hline

CorrNet  & {\bf 91.8} & 74.2 & {\bf 84.6} & {\bf 74.2} & 49.0 & {\bf 64.4} \\
\hline
Klementiev et al.  & 77.6 & 71.1 & 74.5 & 61.9 & 31.3 & 63.0 \\
\hline
MT  & 68.1 & 67.4 & 76.3 & 71.1 & \textbf{52.0} & 58.4 \\
\hline
Majority Class  & 46.8 & 46.8 & 22.5 & 25.0 & 15.3 & 22.2 \\
\hline
Hermann and Blunsom & 88.1 & {\bf 79.1} & N.A. & N.A. & N.A. & N.A. \\
\hline
\end{tabular}
\end{footnotesize}
\end{center}

\caption {Cross-lingual classification accuracy for 3 language pairs, with 1000 labeled examples.} \label{tab:title} 

\vspace*{-0.5cm}
\end{table}

\begin{figure}[H]
\vskip 0.2in
\begin{center}
\includegraphics[height=70mm]{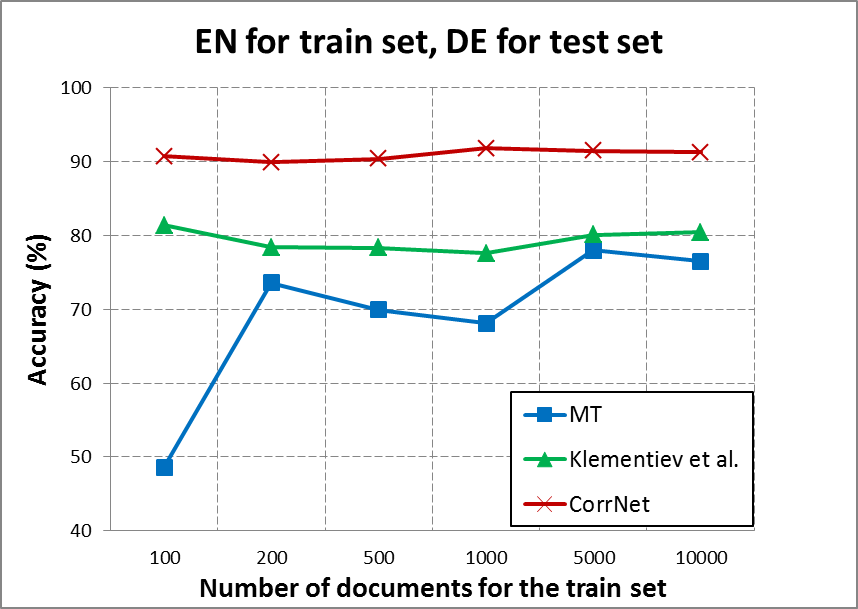}
\caption{Cross-lingual classification accuracy results for EN $\rightarrow$ DE}
\label{ff1}
\end{center}
\vskip -0.2in
\end{figure}

\begin{figure}[H]
\vskip 0.2in
\begin{center}
\includegraphics[height=70mm]{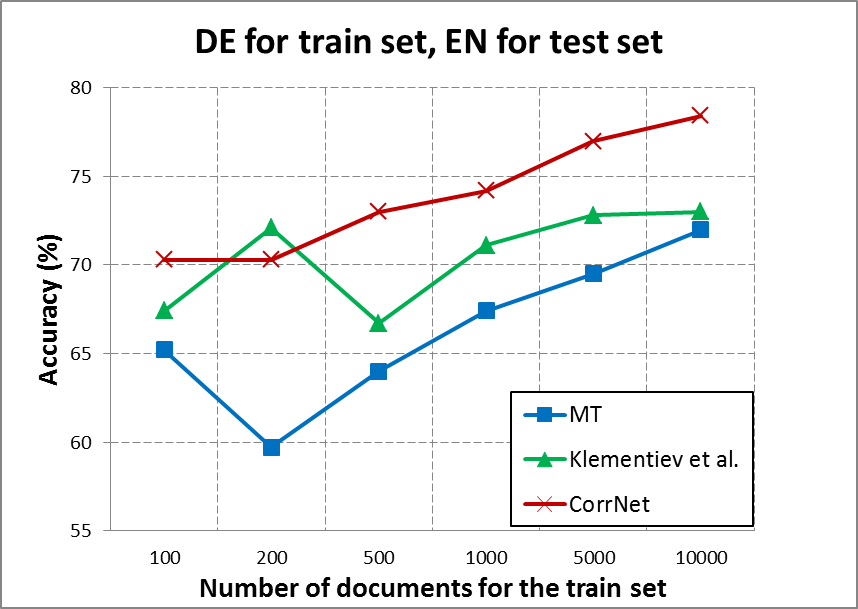}
\caption{Cross-lingual classification accuracy results for DE $\rightarrow$ EN}
\label{ff2}
\end{center}
\vskip -0.2in
\end{figure}

We also evaluate the effect of varying the amount of supervised training data for training the classifier. For brevity, we report only the results for the EN/DE pair, which are summarized in Figure~\ref{ff1} and Figure~\ref{ff2}. We observe that CorrNet clearly outperforms the other models at almost all data sizes. More importantly, it performs remarkably well at very low data sizes (100), suggesting it learns very meaningful embeddings, though the method can still benefit from more labeled data (as in the DE $\rightarrow$ EN case).

\begin{table}
\begin{center}
\begin{footnotesize}
\begin{tabular}{cc|cc|cc}
  
   \multicolumn{2}{c|}{january} & \multicolumn{2}{|c|}{president} & \multicolumn{2}{|c}{said}\\ 
  \hline 
  en & de & en & de & en & de \\
  \hline
  january&januar&president&pr\"asident&said&gesagt\\
march&m\"arz&i&pr\"asidentin&told&sagte\\
october&oktober&mr&pr\"asidenten&say&sehr\\
july&juli&presidents&herr&believe&heute\\
december&dezember&thank&ich&saying&sagen\\
1999&jahres&president-in-office&ratspr\"asident&wish&heutigen\\
june&juni&report&danken&shall&letzte\\
month&1999&voted&danke&again&hier\\
\hline
  \hline

   \multicolumn{2}{c|}{oil} & \multicolumn{2}{|c|}{microsoft} & \multicolumn{2}{|c}{market}\\ 
  \hline 
  en & de & en & de & en & de \\
  \hline
  oil&\"ol&microsoft&microsoft&market&markt\\
supply&boden&cds&cds&markets&marktes\\
supplies&befindet&insider&warner&single&m\"arkte\\
gas&ger\"at&ibm&tageszeitungen&commercial&binnenmarkt\\
fuel&erd\"ol&acquisitions&ibm&competition&m\"arkten\\
mineral&infolge&shareholding&handelskammer&competitive&handel\\
petroleum&abh\"angig&warner&exchange&business&\"offnung\\
crude&folge&online&veranstalter&goods&binnenmarktes\\
  \hline

  \end{tabular}
  \end{footnotesize}
\caption{Example English words along with 8 closest words both in English (en) and German (de), using the Euclidean distance between the embeddings learned by CorrNet}
  
  \label{Tabd}   
  \end{center}
\end{table}

Table~\ref{Tabd} also illustrates the properties captured within and across languages, for the EN/DE pair. For a few English words, the words with closest word representations (in Euclidean distance) are shown, for both English and German. We observe that words that form a translation pair are close, but also that close words within a language are syntactically/semantically similar as well.

The excellent performance of CorrNet suggests that merging several sentences
into single bags-of-words can still yield good word embeddings. In other words, not only we do not need to rely on word-level alignments, but exact sentence-level alignment is also not essential to reach good performances. 
We experimented with the merging of 5, 25 and 50 adjacent sentences into a single bag-of-words. Results are shown in Table~\ref{tab:title2}. They suggest that merging several sentences into single bags-of-words does not necessarily impact the quality of the word embeddings. Thus they confirm that exact sentence-level alignment is not essential to reach good performances as well. 

\begin {table}[H]
\begin{center}
\begin{footnotesize}
\begin{tabular}{|l|l|l|l|l|l|l|l|}
\hline
{\bf } & \# sent. &  EN $\rightarrow$ DE & DE $\rightarrow$ EN & EN $\rightarrow$ FR & FR $\rightarrow$ EN & EN $\rightarrow$ ES & ES $\rightarrow$ EN \\
\hline\hline

\multirow{3}{*}{CorrNet } & 5 & 91.75 & 72.78 & 84.64 &	74.2 &	49.02 &	64.4\\
\cline{2-8}
                         & 25 & 88.0 & 64.5 & 78.1 &	70.02	& 68.3 & 	54.68 \\
\cline{2-8}
                         & 50 & 90.2 & 49.2 & 82.44	 & 75.5	 & 38.2	 & 67.38 \\
\hline
\end{tabular}
\end{footnotesize}
\caption {Cross-lingual classification accuracy for 3 different pairs of languages, when merging the bag-of-words for different numbers of sentences. These results are based on 1000 labeled examples.} \label{tab:title2} 

\end{center}
\end{table}

\section{Transliteration Equivalence}
\label{sec:transEq}

In the previous section, we showed the application of CorrNet in a cross language learning setup. In addition to cross language learning, CorrNet can also be used for matching equivalent items across views. As a case study, we consider the task of determining transliteration equivalence of named entities wherein given a word $u$ written using the script of language $X$ and a word $v$ written using the script of language $Y$ the goal is to determine whether $u$ and $v$ are transliterations of each other. Several approaches have been proposed for this task and the one most related to our work is an approach which uses CCA for determining transliteration equivalence.  

We condider English-Hindi as the language pair for which transliteration equivalence needs to be determined. For learning common representations we used approximately 15,000 transliteration pairs from NEWS 2009 English-Hindi training set \citep{news11}. We represent each Hindi word as a bag of $2860$ bigram characters. This forms the first view ($\xx_i$). Similarly we represent each English word as a bag of $651$ bigram characters. This forms the second view ($\yy_i$). Each such pair $(\xx_i, \yy_i)$ then serves as one training instance for the CorrNet. 

For testing we consider the standard NEWS 2010 transliteration mining test set ~\citep{news10}. This test set contains approximately 1000 Wikipedia English Hindi title pairs. The original task definition is as follows. For a given English title containing $T_1$ words and the corresponding Hindi title containing $T_2$ words identify all pairs which form a transliteration pair. Specifically, for each title pair, consider all $T_1 \times T_2$ word pairs and identify the correct transliteration pairs. In all, the test set contains $5468$ word pairs out of which $982$ are transliteration pairs. For every word pair $(\xx_i, \yy_i)$ we obtain a 50 dimensional common representation for $\xx_i$ and $\yy_i$ using the trained CorrNet. We then calculate the correlation between the representations of $\xx_i$ and $\yy_i$. If the correlation is above a threshold we mark the word pair as equivalent. This threshold is tuned using an additional 1000 pairs which were provided as training data for the NEWS 2010 transliteration mining task. As seen in Table \ref{Tab3} CorrNet clearly performs better than the other methods. Note that our aim is not to achieve state of the art performance on this task but to compare the quality of the shared representations learned using different CRL methods considered in this paper. 

\begin{table}[H]
\begin{center}
\begin{tabular}{|c|c|}
  \hline 
  Model & F1-measure (\%) \\ 
  \hline 
  CCA & 49.68\\
  KCCA & 42.36\\
  MAE & 72.75\\
  CorrNet & \textbf{81.56}\\
 
  \hline
  
  \end{tabular}
  \caption{Performance on NEWS 2010 En-Hi Transliteration Mining Dataset}
  \label{Tab3}   
  \end{center}
\end{table}

\section{Bigram similarity using multilingual word embedding}
\label{sec:bigram}
In this section, we consider one more dataset/application to compare the performance of CorrNet with other state of the art methods. Specifically, the task at hand is to calculate the similarity score between two bigram pairs in English based on their representations. These bigram representations are calculated from word representations learnt using English German word pairs. The motivation here is that the German word provides some context for disambiguating the English word and hence leads to better word representations. This task has been already considered in \cite{mitchell}, \cite{lu} and \cite{wang}. We follow the similar setup as \cite{wang} and use the same dataset. The English and German words are first represented using 640-dimensional monolingual word vectors trained via Latent Semantic Indexing (LSI) on the WMT 2011 monolingual news corpora. We used 36,000 such English-German monolingual word vector pairs for common representation learning. Each pair consisting of one English ($x_i$) and one German($y_i$) word thus acts as one training instance, $z_i = (x_i, y_i)$, for the CorrNet. Once a common representation is learnt, we project all the English words into this common subspace and use these word embeddings for computing similarity of bigram pairs in English.

The bigram similarity dataset was initially used in \cite{mitchell}. We consider the adjective-noun (AN) and verb-object (VN) subsets of the bigram similarity dataset. We use the same tuning and test splits of size 649/1,972 for each subset. The vector representation of a bigram is computed by simply adding the vector representations of the two words in the bigram. Following previous work, we compute the cosine similarity between the two vectors of each bigram pair, order the pairs by similarity, and report the Spearman's correlation ($\rho$) between the model's ranking and human rankings.

Following \cite{wang}, we fix the dimensionality of the vectors at L = 384. Other hyperparameters are tuned using the tuning data. The results are reported in Table \ref{Tab4} where we compare CorrNet with different methods proposed in \cite{wang}. CorrNet performs better than the previous state-of-the-art (DCCAE) on average score. Best results are obtained using CorrNet-500-384. This experiment suggests that apart from multiview applications such as (i) transfer learning (ii) reconstructing missing view and (iii) matching items across views, CorrNet can also be employed to exploit multiview data to improve the performance of a single view task (such as monolingual bigram similarity). 

\begin{table}[H]
\begin{center}
\begin{tabular}{|c|c|c|c|}
  \hline 
  Model & AN & VN & Avg. \\ 
  \hline 
  Baseline (LSI) & 45.0 & 39.1 & 42.1\\
  CCA & 46.6 & 37.7 & 42.2\\
  SplitAE & 47.0 & 45.0 & 46.0\\
  CorrAE & 43.0 & 42.0 & 42.5\\
  DistAE & 43.6 & 39.4 & 41.5\\
  FKCCA & 46.4 & 42.9 & 44.7\\
  NKCCA & 44.3 & 39.5 & 41.9\\
  DCCA & 48.5 & 42.5 & 4.5\\
  DCCAE & \textbf{49.1} & 43.2 & 46.2\\
  CorrNet & 46.2 & \textbf{47.4} & \textbf{46.8}\\
  
  \hline
  
  \end{tabular}
  \caption{Spearman's correlation for bigram similarity dataset. Results for other models are taken from \cite{wang}}
  \label{Tab4}   
  \end{center}
\end{table}

\section{Conclusion and Future Work}
\label{sec:concFutureWork}
In this paper, we proposed Correlational Neural Networks as a method for learning common representations for two views of the data. The proposed model has the capability to reconstruct one view from the other and it ensures that the common representations learned for the two views are aligned and correlated. Its training procedure is also scalable. Further, the model can benefit from additional single view data, which is often available in many real world applications. We employ the common representations learned using CorrNet for two downstream applications, \textit{viz.}, cross language document classification and transliteration equivalence detection. For both these tasks we show that the representations learned using CorrNet perform better than other methods. 

We believe it should be possible to extend CorrNet to multiple views. This could be very useful in applications where varying amounts of data are available in different views. For example, typically it would be easy to find parallel data for English/German and English/Hindi, but harder to find parallel data for German/Hindi. If data from all these languages can be projected to a common subspace then English could act as a pivot language to facilitate cross language learning between Hindi and German. We intend to investigate this direction in future work.

\section*{Acknowledgement}

We would like to thank Alexander Klementiev and Ivan Titov for providing the code for the classifier
and data indices for the cross language document classification task. We would like to thank Janarthanan Rajendran for valuable discussions.

\bibliographystyle{apalike-new}

\end{document}